\pdfoutput=1

\documentclass[11pt]{article}

\usepackage{acl}

\usepackage{times}
\usepackage{latexsym}

\usepackage[T1]{fontenc}

\usepackage[utf8]{inputenc}

\usepackage{microtype}

\usepackage{amsmath}
\usepackage{amssymb}
\usepackage{hyperref}
\usepackage{graphicx,xcolor}
\usepackage{caption}
\usepackage[subrefformat=parens]{subcaption}
\usepackage{enumitem}
\usepackage{here}

\DeclareMathOperator*{\argmax}{arg\,max}

\newcommand{\setbar}{\:|\:}
\newcommand{\eqspace}{\:=\:}

\newcommand{\seq}{s}
\newcommand{\sprefix}[1]{\seq_{0:#1}}
\newcommand{\binlabel}{y}
\newcommand{\binlabels}{y_{0:|\seq|-1}}
\newcommand{\maxlen}{N}
\newcommand{\maxchange}{M}
\newcommand{\numchange}{m}

\newcommand{\model}{\mathcal{M}}
\newcommand{\modelf}[1]{\model(#1)}
\newcommand{\dataset}{\mathcal{D}}
\newcommand{\inputspace}{\mathcal{I}}
\newcommand{\outputspace}{\mathcal{O}}
\newcommand{\stra}{\mathrm{a}}
\newcommand{\strb}{\mathrm{b}}

\newcommand{\interpolationloss}{\mathcal{L}_{\mathrm{CE}}}
\newcommand{\testindex}{\mathcal{T}}
\newcommand{\domfreq}{\omega_{\mathrm{dom}}}
\newcommand{\kmax}{k_{\mathit{max}}}
\newcommand{\dft}[1]{F[#1]}
\newcommand{\imgnum}{\sqrt{-1}}

\newcommand{\hiddensize}{\mathit{hidden\_size}}
\newcommand{\numlayers}{\mathit{num\_layers}}

\title{Empirical Analysis of the Inductive Bias of Recurrent Neural Networks by Discrete Fourier Transform of Output Sequences}

\author{Taiga Ishii \and Ryo Ueda \and Yusuke Miyao \\
  University of Tokyo \\
  \texttt{\{taigarana, ryoryoueda, yusuke\}@is.s.u-tokyo.ac.jp}
}

\begin{document}
\maketitle


\begin{abstract}
A unique feature of Recurrent Neural Networks (RNNs) is that it incrementally processes input sequences.
In this research, we aim to uncover the inherent generalization properties, i.e., inductive bias, of RNNs with respect to how frequently RNNs switch the outputs through time steps in the sequence classification task, which we call output sequence frequency.
%
Previous work analyzed inductive bias by training models with a few synthetic data and comparing the model's generalization with candidate generalization patterns.
%
However, when examining the output sequence frequency, previous methods cannot be directly applied since enumerating candidate patterns is computationally difficult for longer sequences.
To this end, we propose to directly calculate the output sequence frequency for each model by regarding the outputs of the model as discrete-time signals and applying frequency domain analysis.
%
Experimental results showed that Long Short-Term Memory (LSTM) and Gated Recurrent Unit (GRU) have an inductive bias towards lower-frequency patterns, while Elman RNN tends to learn patterns in which the output changes at high frequencies.
We also found that the inductive bias of LSTM and GRU varies with the number of layers and the size of hidden layers.

\end{abstract}

\section{Introduction}
\begin{figure}[t]
    \centering
    \includegraphics[width=7.5cm]{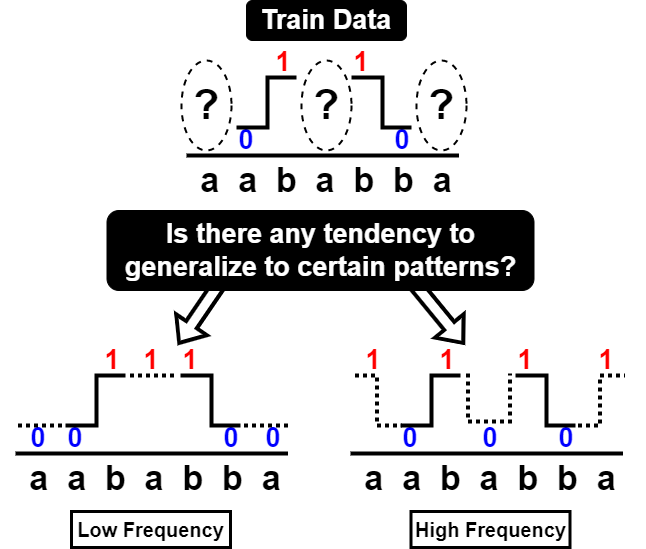}
    \caption{An example showing a train dataset and two candidate generalization patterns, each showing a different output sequence frequency. Here, "aababba" is the input sequence, and there are four binary train labels $0, 1, 1, 0$ each corresponding to the prefix of length $2, 3, 5, 6$.}
    \label{fig:intro}
\end{figure}

In this research, we aim to uncover the inherent generalization properties, i.e., inductive bias, of RNNs with respect to how frequently RNNs switch the outputs through time steps in the sequence classification task, which we call \textit{output sequence frequency} (\autoref{fig:intro}).

In supervised learning settings, a model is trained with finite input-output examples $\{(x_0,\, y_0), \ldots, (x_n,\, y_n) \}$ and then tested with unseen input-output pairs.
The models that achieve high accuracy on test data are often said to ``generalize well''.
However, the important point is that function $f$ that satisfies $f(x_i) = y_i$ cannot be uniquely determined by finite train examples.
This entails that if a model generalizes well to a certain function $f$, then the model hardly generalizes to another function $f'$ that has different outputs for the same unseen inputs, i.e., $f(x_{\mathrm{test}}) \neq f'(x_{\mathrm{test}})$ but is consistent with the same train examples; $f'(x_i) = y_i$.
Therefore, it is crucial to understand what kind of functions a model inherently prefers to learn, which is referred to as \textbf{inductive bias}~\cite{whiteExaminingInductiveBias2021, kharitonov2020WhatTheyWhena,deletang2022NeuralNetworksChomsky, lovering2020PredictingInductiveBiases}.

Our target is Recurrent Neural Network (RNN): a well-known deep learning architecture.
A key feature of RNN is that it processes the input incrementally and predicts the output at each time step, producing a sequence of outputs.
This is different from other deep learning architectures, e.g., Feed Forward Network (FFN), Convolutional Neural Network (CNN), and Transformers~\cite{vaswani2017AttentionAllYou}.
Due to the incremental processing feature of RNNs, the inputs can be of variable length; RNNs have been used for various tasks in natural language processing, such as sentence classification and text generation.
It has also been used as a subcomponent of more complex architectures~\cite{dyer2016RecurrentNeuralNetwork} and to simulate human sequential processing~\cite{steinert-threlkeld2019LearnabilitySemanticUniversals}.
Variants of RNN architectures have been proposed so far.
The most basic one is the Elman RNN~\cite{elman1990FindingStructureTime}.
Later, more complex architectures, such as LSTM~\cite{hochreiter1997LongShortTermMemorya} and GRU~\cite{cho2014LearningPhraseRepresentationsa}, have been proposed to improve modeling long-term dependencies.

Although deep learning models, including RNNs, are said to be high-performance models, they are essentially black boxes, and it is not clear what inductive bias they may have.
In this research, in order to analyze the inductive bias of RNNs, we propose to calculate the output sequence frequency by regarding the outputs of RNNs as discrete-time signals and applying frequency domain analysis.
Specifically, we apply discrete Fourier transform (DFT) to the output signals and compute the dominant frequencies to grasp the overall output patterns.

Inductive bias is not straightforward to analyze since it can be affected by various factors such as the task, dataset, and training method; theoretical analysis has been limited to simple architecture such as FFN~\cite{rahaman2019SpectralBiasNeural,valle-perez2018deep}.
Therefore, empirical studies have been conducted to clarify the inductive bias in various tasks and settings, such as language modeling~\cite{whiteExaminingInductiveBias2021}, sequence classification~\cite{lovering2020PredictingInductiveBiases}, and sequence-to-sequence~\cite{kharitonov2020WhatTheyWhena}.
These works approached the problems by designing synthetic datasets and testing several generalization patterns.
However, when examining the output sequence frequency, we cannot directly apply these previous methods since enumerating exponentially many output sequence patterns in longer sequences is computationally difficult.
To this end, our method makes use of frequency domain analysis to directly calculate the output sequence frequencies and avoid enumerating the candidate generalization patterns.

In the experiment, we randomly generated $500$ synthetic datasets and trained models on a few data points (\autoref{fig:intro}).
As a result, we found:
\begin{itemize}[noitemsep]
    \item LSTM and GRU have an inductive bias such that the output changes at lower frequencies compared to Elman RNN, which can easily learn higher frequency patterns,
    \item The inductive bias of LSTM and GRU varies with the number of layers and the size of hidden layers.
\end{itemize}

\section{Background}

\subsection{Inductive Bias Analysis}\label{subsection:prevworks}
Inductive bias analysis is usually performed by constructing synthetic datasets.
This is because data from real tasks are complex and intertwined with various factors, making it difficult to determine what properties of the dataset affect the behavior of the model.
%
For example, \citet{whiteExaminingInductiveBias2021} targeted LSTM and Transformer and investigated whether easy-to-learn languages differ depending on their typological features in language modeling.
\citet{whiteExaminingInductiveBias2021} used Context Free Grammar (CFG) to construct parallel synthetic language corpora with controlled typological features.
They trained models on each language and computed their perplexities to find that LSTM performs well regardless of word order while the transformer is affected.
%
Another more synthetic example is \citet{kharitonov2020WhatTheyWhena}.
\citet{kharitonov2020WhatTheyWhena} targeted LSTM, CNN, and Transformer.
They designed four synthetic tasks in the sequence-to-sequence framework and trained models on very small datasets (containing 1\textasciitilde 4 data points).
To examine the inductive biases of the models, they prepared a pair of candidate generalization patterns, such as COUNT and MEMORIZATION, for each task and compared the models' preference over the candidate patterns by calculating the Minimum Description Length~\cite{rissanen1978ModelingShortestData}.
Using extremely small train datasets makes it possible to restrict the information models can obtain during training and analyze the models' inherent inductive bias in a more controlled setup.

In this research, we take a similar approach as \cite{kharitonov2020WhatTheyWhena}, restricting the train data to extremely small numbers.
However, we cannot directly apply the methods of \cite{kharitonov2020WhatTheyWhena} because the approach of comparing with candidate generalization patterns can be impractical in our case.
Specifically, when examining the output sequence frequency, it is necessary to feed the models with longer sequences in order to analyze a wide range of frequencies from low to high; there are exponentially many patterns with the same number of output changes in longer sequences, which makes it difficult to exhaustively enumerate the candidate generalization patterns.
Therefore, instead of preparing candidate generalization patterns, we directly calculate the output sequence frequency for each model by regarding the outputs of the model as discrete-time signals and applying frequency domain analysis.

\subsection{Frequency Domain Analysis}
%
Discrete Fourier Transform (DFT) is a fundamental analysis technique in digital signal processing.
Intuitively, DFT decomposes a signal into a sum of finite sine waves of different frequencies, allowing one to analyze what frequency components the original signal consists of.
The DFT for a length $\maxlen$ discrete-time signal $f[0], \ldots, f[\maxlen - 1]$ is defined by the following equation:
\begin{align}\label{eq:dft}
    \dft{k} \eqspace \sum_{n=0}^{\maxlen - 1} f[n]\exp\left(-\imgnum\frac{2\pi}{\maxlen}kn\right).
\end{align}
When $f[n]$ is a real-value signal, it is sufficient to consider only $k\in \{ 1, \ldots, \frac{\maxlen}{2} \}$.\footnote{This is due to the periodicity of $\exp(-\imgnum\frac{2\pi}{\maxlen}kn)$. Furthermore, we do not take into account the $k=0$ term since it is called a DC term and works as an offset. }
Here, $k=1$ corresponds to the lowest frequency component and $k = \frac{\maxlen}{2}$ to the highest.

One useful measure for analyzing the property of the signal $f[n]$ is the dominant frequency~\cite{ng2007UnderstandingInterpretingDominant}.
In short, dominant frequency is the frequency component of maximum amplitude and is expected to represent the general periodic pattern of the original signal $f[n]$.
The dominant frequency $\domfreq$ is defined by $\domfreq \eqspace \frac{2\pi}{N}\kmax$, where $\kmax \eqspace \argmax\{ |\dft{k}| \}$.

\section{Methods}

\subsection{Task}
To analyze the output sequence frequency, i.e., how frequently the output changes through time steps, we focus on a simple case of binary sequence classification task: the inputs are the prefixes of a binary sequence $\seq \in \{\stra, \strb\}^*$.
Specifically, given a binary sequence $\seq \in \{\stra, \strb\}^*$, the input space $\inputspace$ and the output space $\outputspace$ are defined as follows:
\begin{align}
    \inputspace &\eqspace \{ \sprefix{i} \setbar i = 0, \ldots |\seq|-1 \},\\
    \outputspace &\eqspace \{(1-p, p) \setbar p \in [0, 1]\},
\end{align}
where $\outputspace$ is a set of categorical distributions over the binary labels $\{0, 1\}$, and $p$ denotes the probability of predicting label $1$.

Without loss of generality, we can only consider the model's output probability of predicting label $1$ for the sequence $\sprefix{i}$, which we denote by $\modelf{\sprefix{i}}$.
In this way, we can regard the model's output sequence $\modelf{\sprefix{0}}, \ldots, \modelf{\sprefix{|\seq|-1}}$ as a discrete-time signal taking values in $[0, 1]$.

\subsection{Train Dataset}
\label{subsection:train_data}
\begin{figure}
    \centering
    \includegraphics[width=7.5cm]{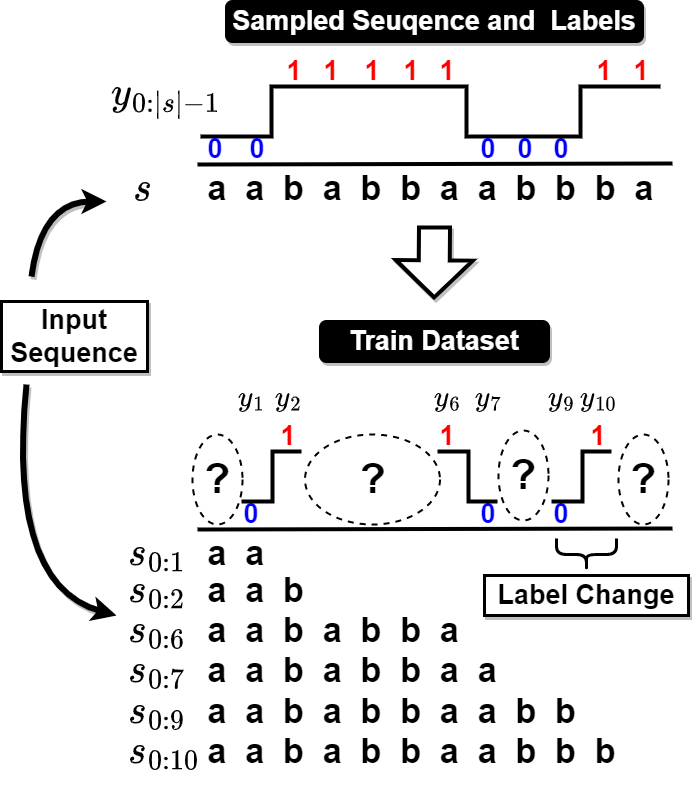}
    \caption{Illustration of train dataset construction. The train dataset contains only the instances corresponding to the label changes.}
    \label{fig:dataset}
\end{figure}
\autoref{fig:dataset} shows an intuitive illustration of our dataset construction.
Given a sequence $\seq$, we randomly generate the binary labels $\binlabels$, where each $\binlabel_i$ is the label assigned to the prefix $\sprefix{i}$.
When two successive labels $\binlabel_i$ and $\binlabel_{i+1}$ differ, we say there is a \textit{label change} (e.g., $\binlabel_{9}$ and $\binlabel_{10}$ in \autoref{fig:dataset}).\footnote{Similarly, we use \textit{output change} for output sequences.}
We then make a train dataset $\dataset$ by taking instances where the labels change: $\{ (\sprefix{i}, \binlabel_i), (\sprefix{i+1}, \binlabel_{i+1}) \setbar \binlabel_{i} \neq \binlabel_{i+1}\}$.
For example, in \autoref{fig:dataset}, the train data $\dataset$ contains $\{ (\stra \stra, 0)\, (\stra \stra \strb, 1)\, (\stra \stra \strb \stra \strb \strb \stra, 1)\, (\stra \stra \strb \stra \strb \strb \stra \stra, 0), \ldots \}$.
Note that the original labels $\binlabels$ can be uniquely recovered from $\dataset$ simply by \textit{interpolating} or \textit{extending} the labels for other prefixes.

The procedure is formalized as follows:
\begin{enumerate}
    \item Sample a sequence $\seq \in \{0, 1\}^\maxlen$, where $\maxlen$ is the length of the sequence,
    \item Sample the number of label changes $\numchange \in \{1, \ldots \maxchange \}$, where $\maxchange$ is the maximum number of label changes,
    \item Sample the labels $\binlabels$ so that all the $\numchange$ label changes do not overlap\footnote{This condition ensures that the labels in the train dataset are balanced.}, i.e. $\forall i, j. \; i < j \wedge \binlabel_{i} \neq \binlabel_{i+1} \wedge \binlabel_{j} \neq \binlabel_{j+1} \Rightarrow i + 1 < j$,
    \item Create a dataset as \\$ \dataset \eqspace \{ (\sprefix{i}, \binlabel_i), (\sprefix{i+1}, \binlabel_{i+1}) \setbar \binlabel_{i} \neq \binlabel_{i+1}\}$.
\end{enumerate}
By training models on random input sequences $\seq$, we expect the model predictions to represent the inherent generalization property of the model.

\subsection{Evaluation Metrics}
For the analysis, we apply two evaluation metrics.

\subsubsection{Test Cross-entropy Loss}
First, we compare the model's output sequence $\modelf{\sprefix{0}}, \ldots, \modelf{\sprefix{|\seq|-1}}$ with the original labels $\binlabels$ by calculating test cross-entropy loss $\interpolationloss$.
Intuitively, near-zero $\interpolationloss$ indicates that the model generalizes to simply \textit{interpolate} or \textit{extend} the training labels since we constructed the train datasets so that the original labels can be recovered by interpolation, as described in \autoref{subsection:train_data}.

The loss is formalized as:
\begin{align}
    \begin{split}
        \interpolationloss \eqspace - \frac{1}{|\testindex|}\sum_{i \in \testindex }  ( & \binlabel_{i} \ln (\modelf{\sprefix{i}}) \\
        + (1- & \binlabel_{i}) \ln (1- \modelf{\sprefix{i}}) ),
    \end{split}
\end{align}
where $\testindex = \{ i \setbar (\sprefix{i}, \_) \notin \dataset \}$ is the set of test data indices.

\subsubsection{Dominant Frequency}
In case $\interpolationloss$ is high, we consider the model's output sequence $\modelf{\sprefix{0}}, \ldots, \modelf{\sprefix{|\seq|-1}}$ as a discrete-time signal and apply frequency domain analysis to look into the model's behavior.
More specifically, we apply DFT to the output signal and obtain the dominant frequency $\domfreq$.
The dominant frequency $\domfreq$ is calculated by simply replacing $f[n]$ in \autoref{eq:dft} with $\modelf{\sprefix{n}}$.

\subsection{Experiment Settings}\label{subsection:settings}
Here, we describe the basic settings of our experiment.
We use well-known basic RNN architectures: LSTM~\cite{hochreiter1997LongShortTermMemorya}, GRU~\cite{cho2014LearningPhraseRepresentationsa}, and Elman RNN~\cite{elman1990FindingStructureTime}.
For the decoding, we use a linear decoder without bias followed by a softmax function.
We try 4 combinations of hyperparameters: $(\numlayers,\; \hiddensize) \in \{(1, 200), \; (2, 200), \; (3, 200),\; (2, 2000)\}$, where $\numlayers$ denotes the number of layers, and $\hiddensize$ denotes the size of hidden layers.\footnote{For other hyperparameters and parameter initialization, we used the default settings of PyTorch \url{https://pytorch.org/}.}

For optimization, we train models to minimize the average cross-entropy loss by gradient descent using Adam~\cite{kingma2015AdamMethodStochastic} with a learning rate of $1.0\times10^{-4}$ for $1000$ epochs.\footnote{Since the maximum size of train data is 10 in our settings, all the data are put in a batch during the training.}

Finally, we randomly generate $500$ train datasets with $\maxlen \eqspace 100, \maxchange \eqspace 5$ and train $10$ models with different random seeds for each dataset, architecture, and parameter setting.
Note that this sparse setting ($10:90$ train-test data ratio at maximum) keeps the hypothesis space large and thus enables us to analyze the inductive bias of the models as described in \autoref{subsection:prevworks}.

Training all the models took around 30 hours using 8 NVIDIA A100 GPUs.

\section{Findings}

\subsection{Models Do Not Learn to Interpolate}
\begin{figure}[t]
    \centering
    \includegraphics[width=\linewidth]{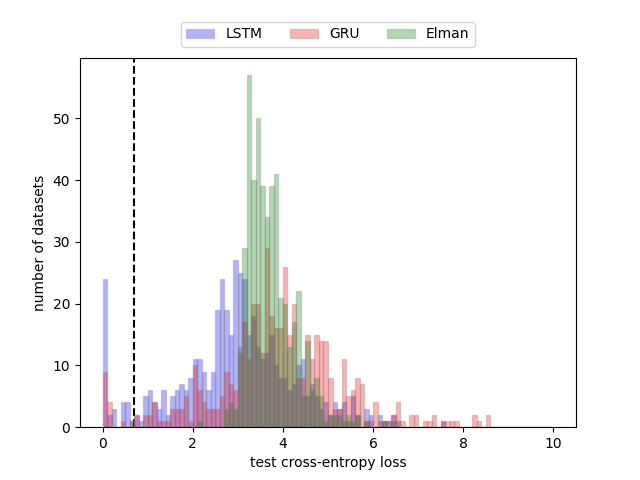}
    \caption{The median test cross-entropy loss counts for LSTM, GRU, and Elman RNN with $(\numlayers,\; \hiddensize)\eqspace (2, 200)$. The dotted vertical line shows the random baseline loss of $-\ln (\frac{1}{2})$.}
    \label{fig:loss}
\end{figure}
\begin{figure*}[t]
    \begin{tabular}{cc}
        \begin{minipage}[t]{\columnwidth}
            \centering
            \includegraphics[width=\columnwidth]{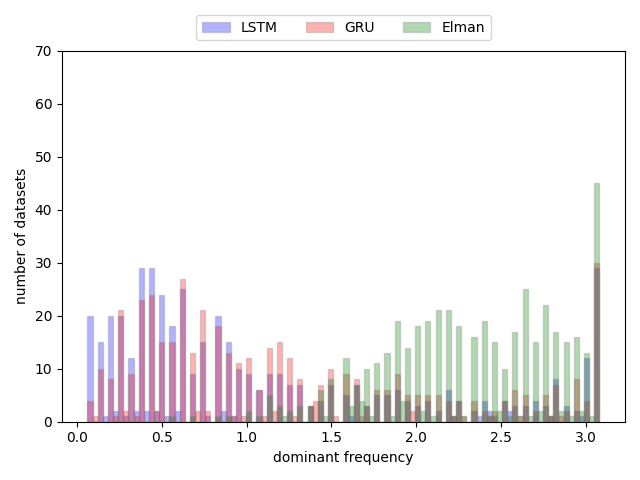}
            \subcaption{The results for $\numlayers=1,\, \hiddensize=200$.}
        \end{minipage} &
        \begin{minipage}[t]{\columnwidth}
            \centering
            \includegraphics[width=\columnwidth]{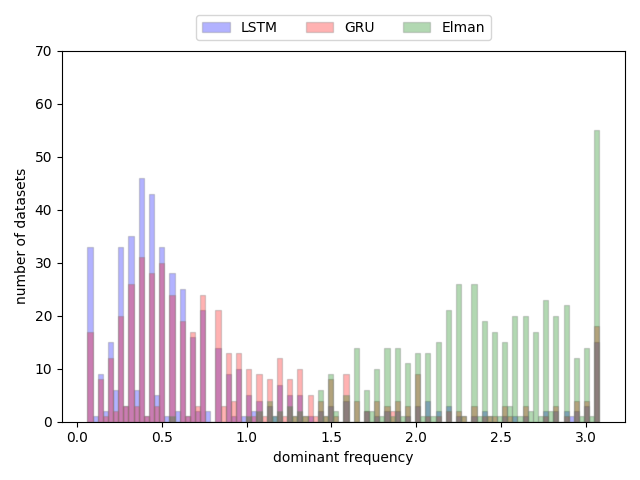}
            \subcaption{The results for our base case $\numlayers=2,\, \hiddensize=200$.}
        \end{minipage} \\
        \begin{minipage}[t]{\columnwidth}
            \centering
            \includegraphics[width=\columnwidth]{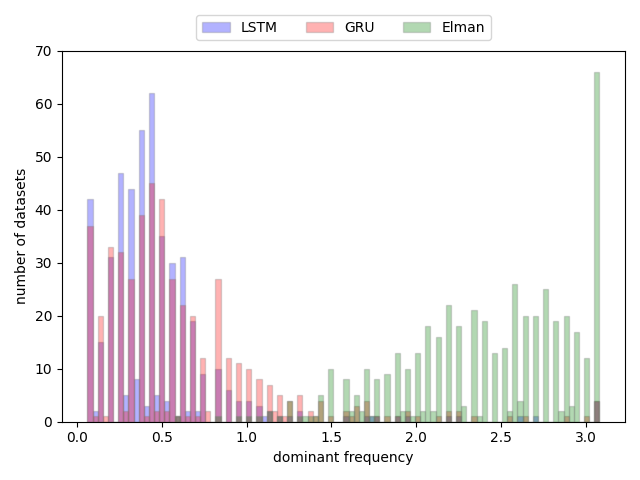}
            \subcaption{The results for $\numlayers=3,\, \hiddensize=200$.}
        \end{minipage} &
        \begin{minipage}[t]{\columnwidth}
            \centering
            \includegraphics[width=\columnwidth]{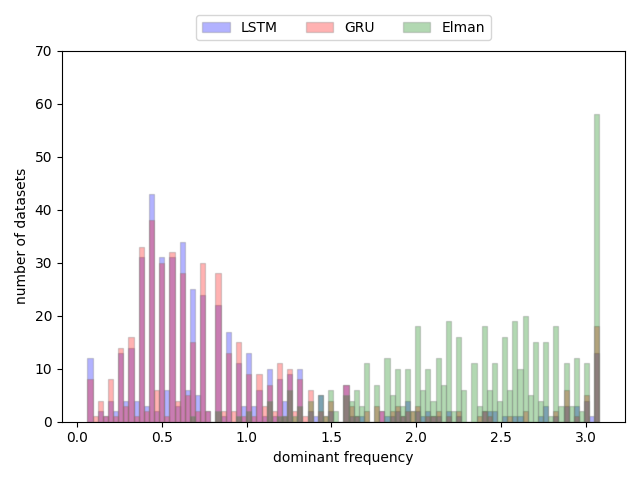}
            \subcaption{The results for $\numlayers=2,\, \hiddensize=2000$.}
        \end{minipage}
    \end{tabular}
    \caption{The median dominant frequency counts for LSTM, GRU, and Elman RNN with different hyperparameters.}
    \label{fig:all_freq}
\end{figure*}
In order to see if the models generalize simply to interpolate the given labels, we calculate the median test cross-entropy loss of the multiple models trained for each dataset (\autoref{fig:loss}).
The dotted vertical line shows the random baseline loss of $-\ln (\frac{1}{2}) \approx 0.7$.
As can be seen in \autoref{fig:loss}, the median test cross-entropy loss is higher than the random baseline for most datasets for all of LSTM, GRU, and Elman RNN.
This indicates that, in most cases, none of the LSTM, GRU, or Elman RNN learns to interpolate in this extremely simple setup, where only the label-changing part is given as training data.
We also observe a similar trend in other hyperparameter settings; The test cross-entropy losses for other settings are shown in \autoref{appendix:loss}.

\subsection{Architectural Difference}
\begin{figure*}[t]
    \begin{tabular}{cc}
      \begin{minipage}[t]{0.48\hsize}
        \centering
        \includegraphics[keepaspectratio, width=\linewidth]{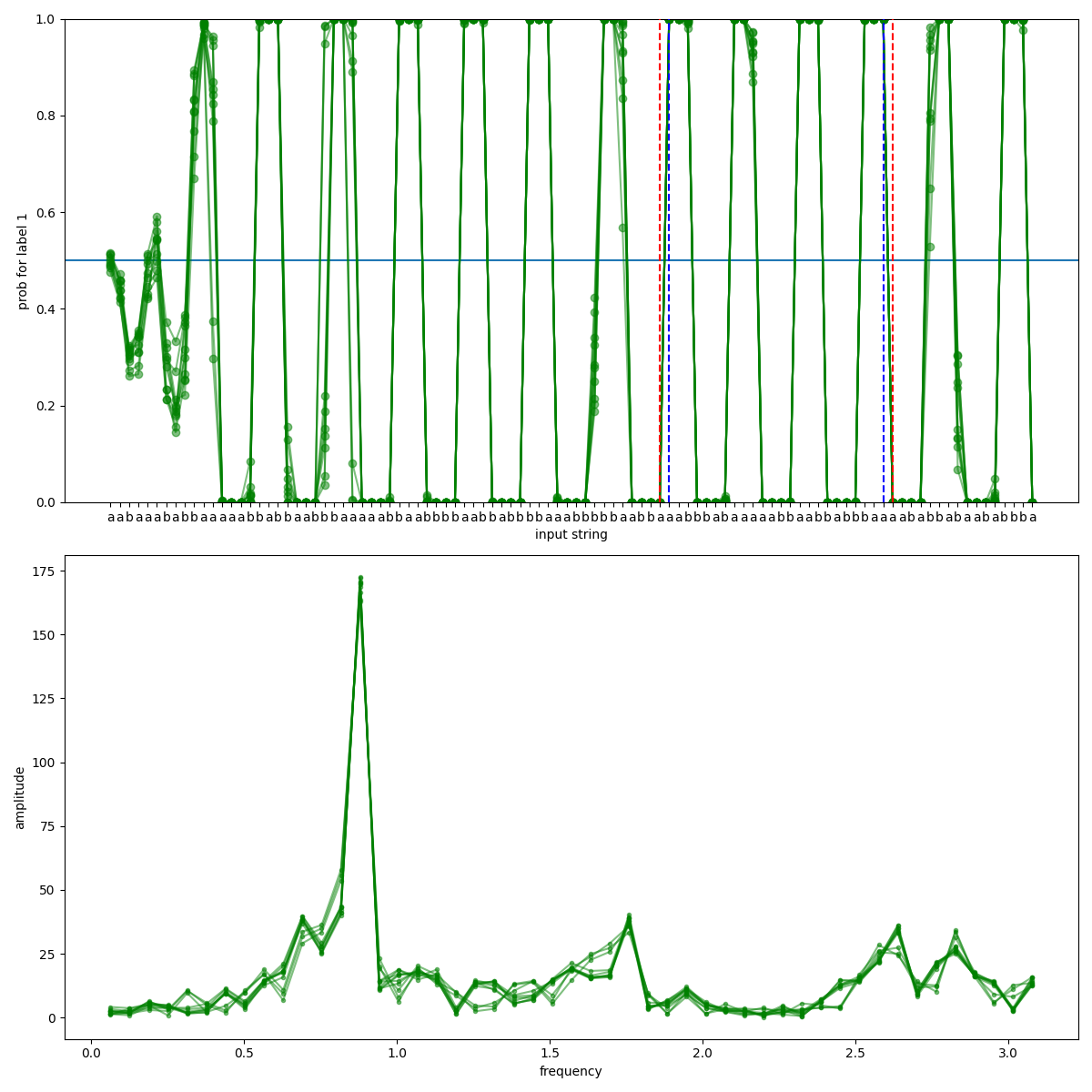}
        \subcaption{LSTM}
        \label{fig:lstm_str}
      \end{minipage} &
      \begin{minipage}[t]{0.48\hsize}
        \centering
        \includegraphics[keepaspectratio, width=\linewidth]{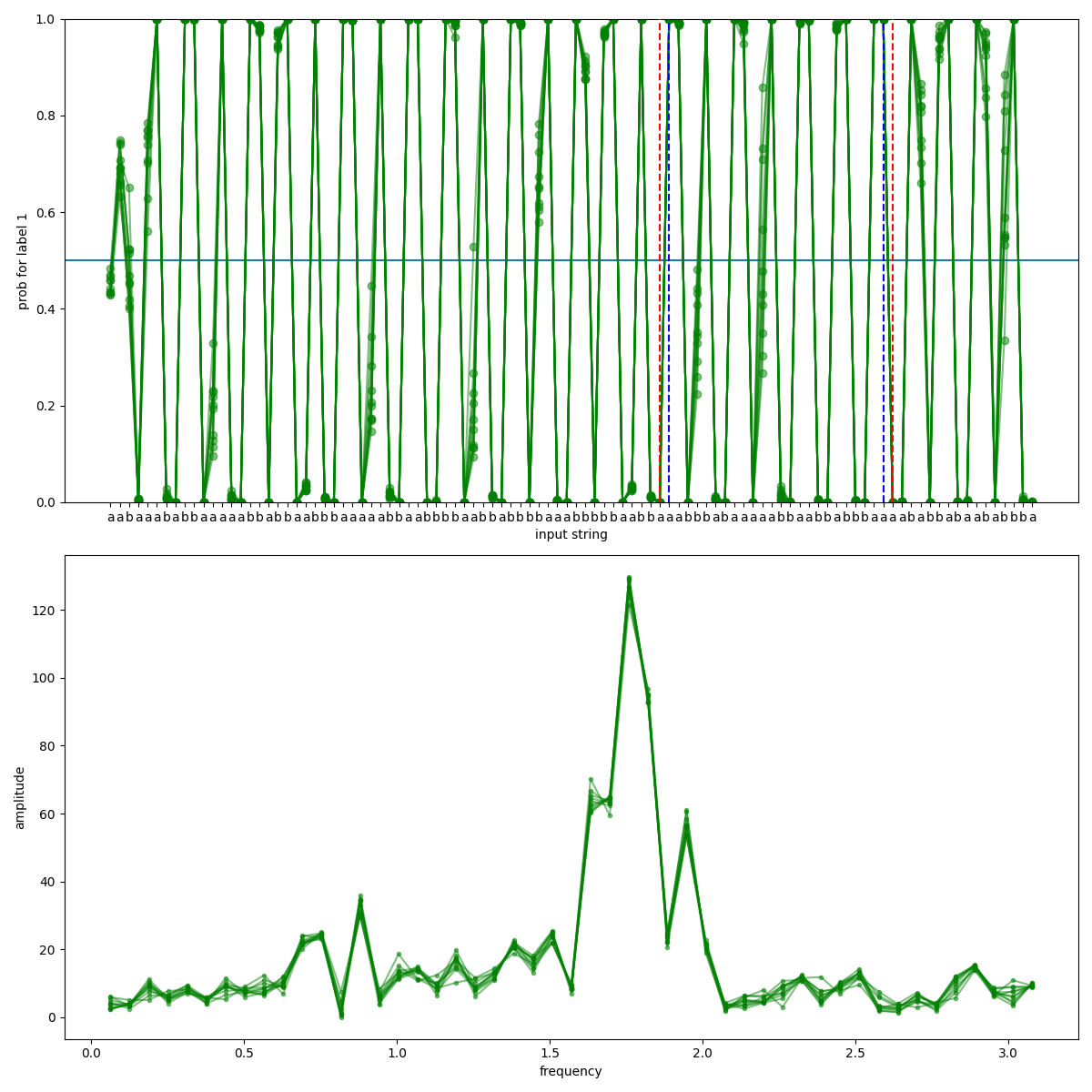}
        \subcaption{Elman RNN}
        \label{fig:elman_str}
      \end{minipage}
    \end{tabular}
    \caption{An example of LSTM and Elman RNN with $(\numlayers,\; \hiddensize)\eqspace (2, 200)$. The top rows show the actual model outputs for a specific sequence, and the bottom rows show the DFT of model outputs. In this example, 4 labels $0, 1, 1, 0$ are assigned to the prefixes of length $60, 61, 84, 85$. The Red and blue vertical lines correspond to the labels $0, 1$, respectively. The results of 10 models with different random seeds are shown.}
    \label{fig:str_sample}
  \end{figure*}
Now that the test cross-entropy loss has revealed that the patterns learned by the models contain more output changes than the original pattern in the train data, the next step is to see if there are any architecture-specific trends in the output sequence patterns.
We calculate the dominant frequency for each model and take the median over the models trained on the same dataset.
\autoref{fig:all_freq} shows the distribution of median dominant frequencies for LSTM, GRU, and Elman RNN with different hyperparameters.
It is clear that, in all settings, LSTM and GRU tend to learn lower-frequency patterns, while the dominant frequencies of Elman RNN tend to be higher.
Comparing LSTM and GRU, LSTM has slightly lower-frequency patterns for $\hiddensize \eqspace 200 $ (\autoref{fig:all_freq} (a, b, c)), though the difference is not as clear for $\hiddensize \eqspace 2000$ (\autoref{fig:all_freq} (d)).

An example of sequential outputs of LSTM and Elman is shown in \autoref{fig:str_sample}.
The top rows show the actual model outputs for a specific sequence, and the bottom rows show the DFT of model outputs.
In this example, only 4 labels $0, 1, 1, 0$ are given to the prefixes of length $60, 61, 84, 85$.
It is clear that both LSTM and Elman learn periodic patterns but do not learn to interpolate the given train labels.
Besides, it is also notable that LSTMs indeed learn lower-frequency patterns compared to Elman RNNs.

\subsection{Effect of Hyperparameters}
\label{subsec:hypera}
Here, we describe how hyperparameters affect the observed inductive biases.

\subsubsection{Number of Layers}
\begin{figure}[p]
    \begin{tabular}{c}
      \begin{minipage}[t]{\linewidth}
        \centering
        \includegraphics[clip, keepaspectratio, width=\linewidth]{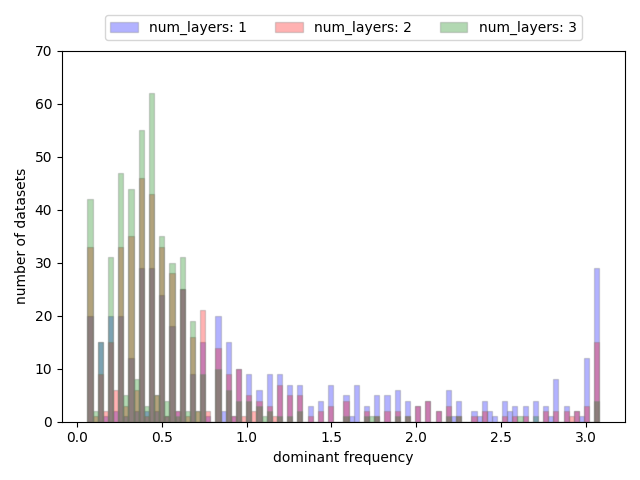}
        \subcaption{LSTM}
        \label{fig:lstm_numlayer}
      \end{minipage} \\
      \begin{minipage}[t]{\linewidth}
        \centering
        \includegraphics[clip, keepaspectratio, width=\linewidth]{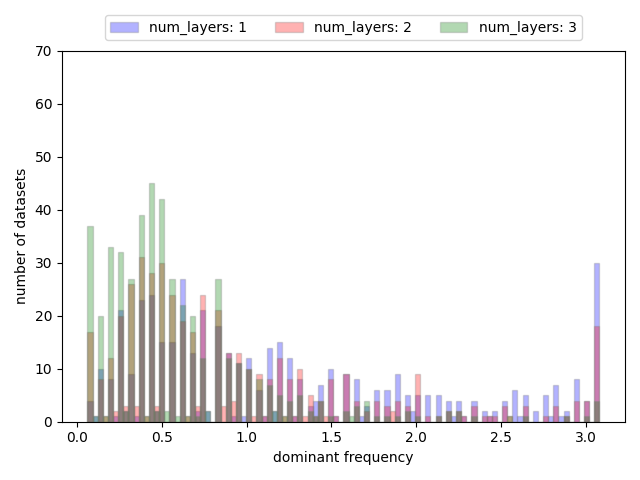}
        \subcaption{GRU}
        \label{fig:gru_num_layer}
      \end{minipage} \\
      \begin{minipage}[t]{\linewidth}
        \centering
        \includegraphics[clip, keepaspectratio, width=\linewidth]{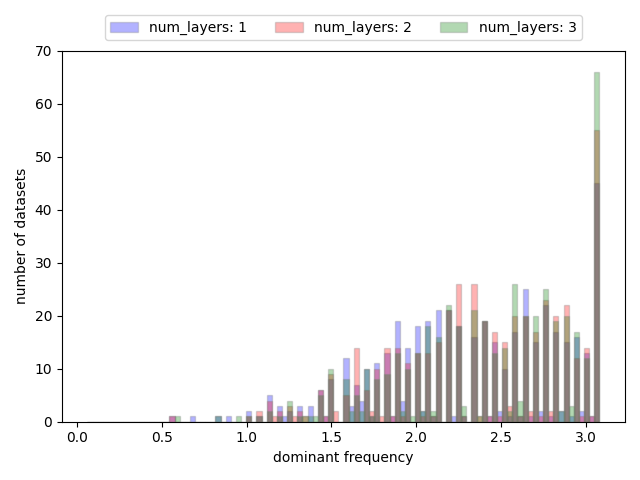}
        \subcaption{Elman RNN}
        \label{fig:elman_num_layer}
      \end{minipage}
    \end{tabular}
    \caption{The median dominant frequencies of $\numlayers=1, 2, 3$ for LSTM, GRU, and Elman RNN with $\hiddensize=200$.}
    \label{fig:num_layers}
\end{figure}
\autoref{fig:num_layers} shows the median dominant frequencies of $\numlayers=1, 2, 3$ for LSTM, GRU, and Elman RNN.
As for LSTM, it can be seen that the proportion of patterns in the lower-frequency domain tends to increase as the number of layers increases.
In other words, despite the increased complexity of the models, LSTMs tend to learn simpler patterns (in the sense that the output changes less).
A similar trend is observed for GRU, although not as clear as for LSTM. 
On the other hand, Elman RNN does not show such apparent differences.

\subsubsection{Hidden Layer Size}
\begin{figure}[p]
    \begin{tabular}{c}
      \begin{minipage}[t]{\columnwidth}
        \centering
        \includegraphics[clip, width=\columnwidth]{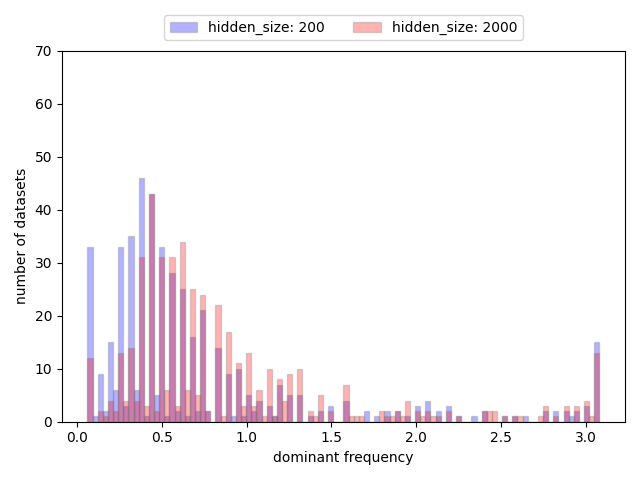}
        \subcaption{LSTM}
        \label{fig:lstm_hidden_size}
      \end{minipage} \\
      \begin{minipage}[t]{\columnwidth}
        \centering
        \includegraphics[clip, width=\columnwidth]{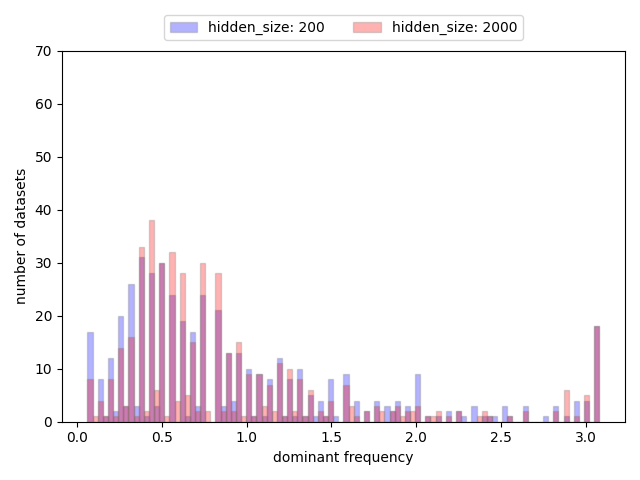}
        \subcaption{GRU}
        \label{fig:gru_hidden_size}
      \end{minipage} \\
      \begin{minipage}[t]{\columnwidth}
        \centering
        \includegraphics[clip, width=\columnwidth]{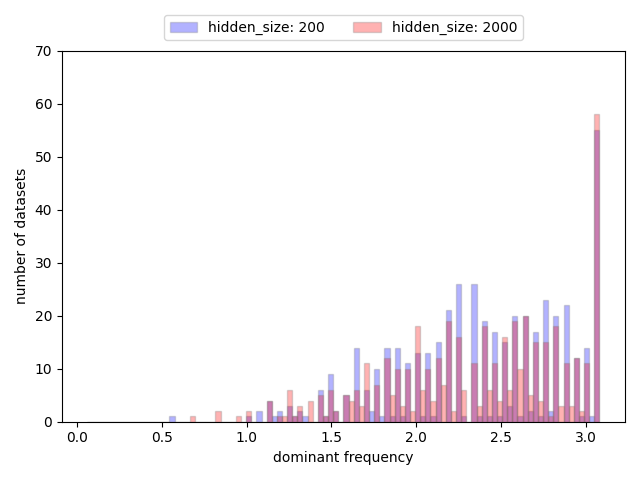}
        \subcaption{Elman RNN}
        \label{fig:elman_hidden_size}
      \end{minipage}
    \end{tabular}
    \caption{The median dominant frequencies of $\hiddensize=200, 2000$ for LSTM, GRU, and Elman RNN with $\numlayers = 2$.}
    \label{fig:hidden_size}
\end{figure}
\autoref{fig:hidden_size} shows the median dominant frequencies of $\hiddensize=200, 2000$ for LSTM, GRU, and Elman RNN.
Although the trend is not so clear, for LSTM and GRU, the counts are slightly larger for $\domfreq = 0.5 \sim 1.0$ when $\hiddensize=2000$, while the counts are larger for $\domfreq=0.0 \sim 0.5$ when $\hiddensize=200$.
This is rather the opposite trend from that of $\numlayers$.
However, the above trend does not seem to appear in Elman RNN.

\section{Discussion and Limitation}

\subsection{Expressive Capacity and Output Sequence Frequency}
Our results do not align with the expressive capacity of RNNs reported in previous work~\cite{merrillFormalHierarchyRNN2020a, weiss2018PracticalComputationalPower}.
\citet{merrillFormalHierarchyRNN2020a,weiss2018PracticalComputationalPower} formally showed that LSTM is strictly more expressive than GRU and Elman RNN.
On the other hand, in our experiments, LSTM and GRU show a bias toward lower frequencies, while Elman RNN, which has the same expressive capacity as GRU, according to \cite{merrillFormalHierarchyRNN2020a}, shows an opposite bias toward higher frequencies.
Note that the expressive capacity and the inductive bias of a model are basically different concepts.
This is because expressive capacity is the theoretical upper bound on the functions a model can represent with all possible combinations of its parameters, regardless of the training procedure.
In contrast, inductive bias is the preference of functions that a model learns from finite train data, possibly depending on training settings.
However, they are not entirely unrelated because a function that is impossible to learn in terms of expressive capacity will never be learned, which can emerge as inductive bias.
We conjecture that the difference between the expressive capacity and the observed inductive bias is due to the simplicity of our experiment setting.
This difference is not a negative result: It indicates that inductive bias in such a simple setting is effective in observing detailed differences that cannot be captured by expressive capacity.

\subsection{Randomness of Outputs}
Previous study showed that FFNs hardly learn random functions since they are inherently biased toward simple structured functions~\cite{valle-perez2018deep}.
We can find a similar trend for RNNs in our experimental results.
In other words, by regarding the outputs of RNNs as discrete-time signals, we can confirm that the signals are not random, i.e., white noises.
If we assume that the output signals of the RNNs are random, the dominant frequency should be uniformly distributed from low to high-frequency regions.
Therefore, the biased distribution in \autoref{fig:all_freq} indicates that the outputs of the RNNs are not random signals.
This is also clear from the example outputs in \autoref{fig:str_sample}, where the models show periodic patterns.

\subsection{Practical Implication}

For LSTM and GRU, we observed different inductive biases between increasing the number of layers and hidden layer size.
Previous study that investigated whether RNNs can learn parenthesis also reported that LSTM and GRU behaved differently when the number of layers and the hidden layer size were increased~\cite{bernardy2018CanRecurrentNeural}.
Although the tasks are different, our findings align with the previous work.
From a practical point of view, these findings suggest that it may be more effective to increase the number of layers than to increase the hidden layer size depending on the target task.

Besides, the fact that LSTM and GRU, which are known to be ``more practical'' than Elman RNN, tend to learn lower frequency patterns may support the idea that output sequence frequency aligns with ``practical usefulness.''
Furthermore, a concept similar to output sequence frequency has been proposed as a complexity measure in sequence classification: sensitivity~\cite{hahn2021SensitivityComplexityMeasure}.
While output sequence frequency focuses on the change in output over string length, sensitivity focuses on the change in output when a string is partially replaced, keeping its length.
It would be an interesting future direction to examine the validity of inductive biases in output sequence frequency as an indicator of complexity and practical usefulness.

\subsection{Limitation}
There are some dissimilarities between our experimental setup and practical sequence classification tasks:
\begin{itemize}[noitemsep]
    \item The task is limited to the binary classification of binary sequences,
    \item Models are trained only on prefixes of a sequence,
    \item The number of train data is extremely small.
\end{itemize}
Therefore, in order to accurately estimate the impact of our findings on the actual task, it is necessary to expand from sequence to language in a multi-label setting with a larger vocabulary.

Due to the computational complexity, we only tried 4 combinations of hyperparameters.
However, it is still necessary to exhaustively try combinations of hyperparameters for a more detailed analysis.

\section{Conclusion}
This study focuses on inductive bias regarding the output sequence frequency of RNNs, i.e., how often RNNs tend to change the outputs through time steps.
To this end, we constructed synthetic datasets and applied frequency domain analysis by regarding the model outputs as discrete-time signals.

Experimental results showed that LSTM and GRU have inductive biases towards having low output sequence frequency, whereas Elman RNN tends to learn higher-frequency patterns.
Such differences in inductive bias could not be captured by the expressive capacity of each architecture alone.
This indicates that inductive bias analysis on synthetic datasets is an effective method for studying model behaviors.

By testing different hyperparameters, we found that the inductive biases of LSTM and GRU vary with the number of layers and the hidden layer size in different ways.
This confirms that when increasing the total number of parameters in a model, it would be effective not only to increase the hidden layer size but also to try various hyperparameters, such as the number of layers.

Although the experimental setting was limited to simple cases, we believe this research shed some light on the inherent generalization properties of RNNs and built the basis for architecture selection and design.


\clearpage

\bibliography{main}
\bibliographystyle{acl_natbib}

\clearpage
\appendix



\begin{figure*}[t]
    \begin{tabular}{cc}
        \begin{minipage}[t]{\columnwidth}
            \centering
            \includegraphics[width=\columnwidth]{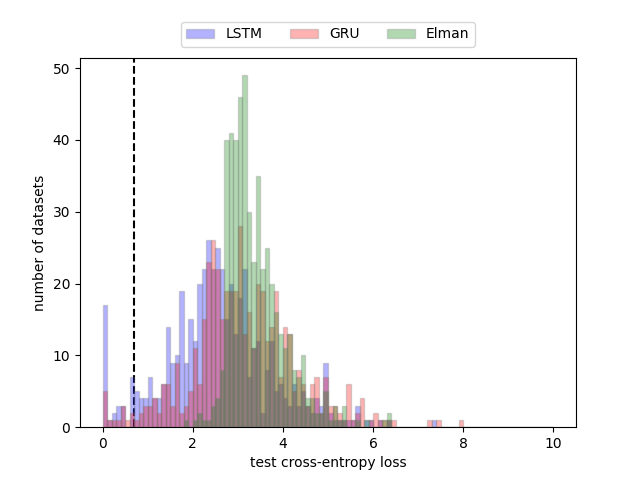}
            \subcaption{The results for $\numlayers=1,\, \hiddensize=200$. In this setting, the mean of the median train cross-entropy loss was at most $0.04$, and the standard deviation was at most $0.09$.}
        \end{minipage} &
        \begin{minipage}[t]{\columnwidth}
            \centering
            \includegraphics[width=\columnwidth]{results/Archi_200_2_0.0-Len_100_Change_5_test.png}
            \subcaption{The results for our base case $\numlayers=2,\, \hiddensize=200$. In this setting, the mean of the median train cross-entropy loss was at most $0.01$, and the standard deviation was at most $0.05$.}
        \end{minipage} \\
        \begin{minipage}[t]{\columnwidth}
            \centering
            \includegraphics[width=\columnwidth]{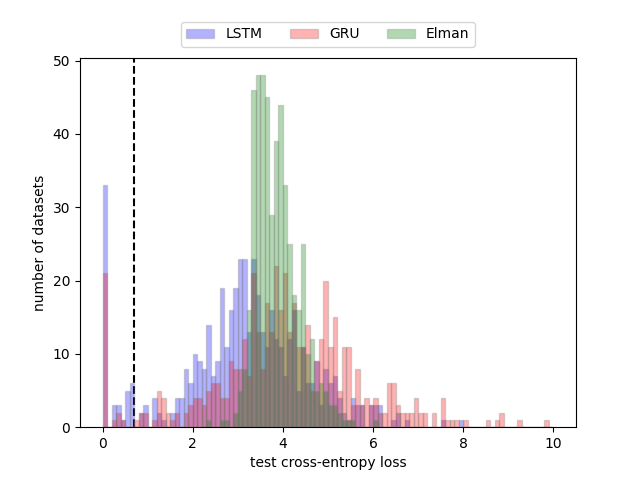}
            \subcaption{The results for $\numlayers=3,\, \hiddensize=200$. In this setting, the mean of the median train cross-entropy loss was at most $0.004$, and the standard deviation was at most $0.03$.}
        \end{minipage} &
        \begin{minipage}[t]{\columnwidth}
            \centering
            \includegraphics[width=\columnwidth]{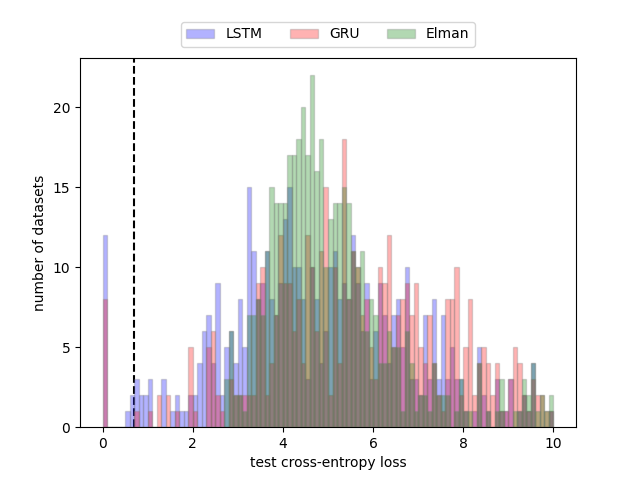}
            \subcaption{The results for $\numlayers=2,\, \hiddensize=2000$. In this setting, the mean of the median train cross-entropy loss was at most $0.02$, and the standard deviation was at most $0.09$.}
        \end{minipage}
    \end{tabular}
    \caption{The median test cross-entropy loss counts for LSTM, GRU, and Elman RNN with different hyperparameters. The dotted vertical line shows the random baseline loss of $-\ln (\frac{1}{2}) \approx 0.7$.}
    \label{fig:all_loss}
\end{figure*}
\section{Test Cross-entropy Loss}\label{appendix:loss}
\autoref{fig:all_loss} shows the distributions of median test cross-entropies in all settings we tried.
As we can see in \autoref{fig:all_loss}, the median test cross-entropy loss is higher than the random baseline for most datasets in all cases.

\section{Raw Cross-entropy Loss}\label{appendix:raw_loss}
\begin{figure*}[t]
    \begin{tabular}{ccc}
        \begin{minipage}[t]{0.3\linewidth}
            \centering
            \includegraphics[width=\linewidth]{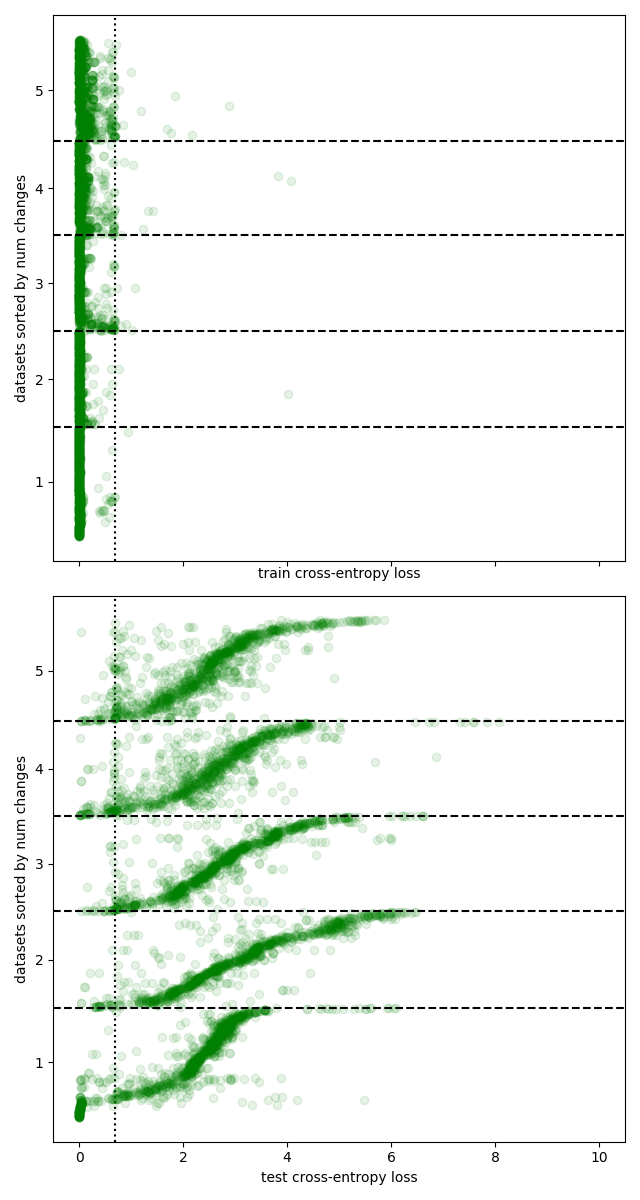}
            \subcaption{LSTM}
        \end{minipage} &
        \begin{minipage}[t]{0.3\linewidth}
            \centering
            \includegraphics[width=\linewidth]{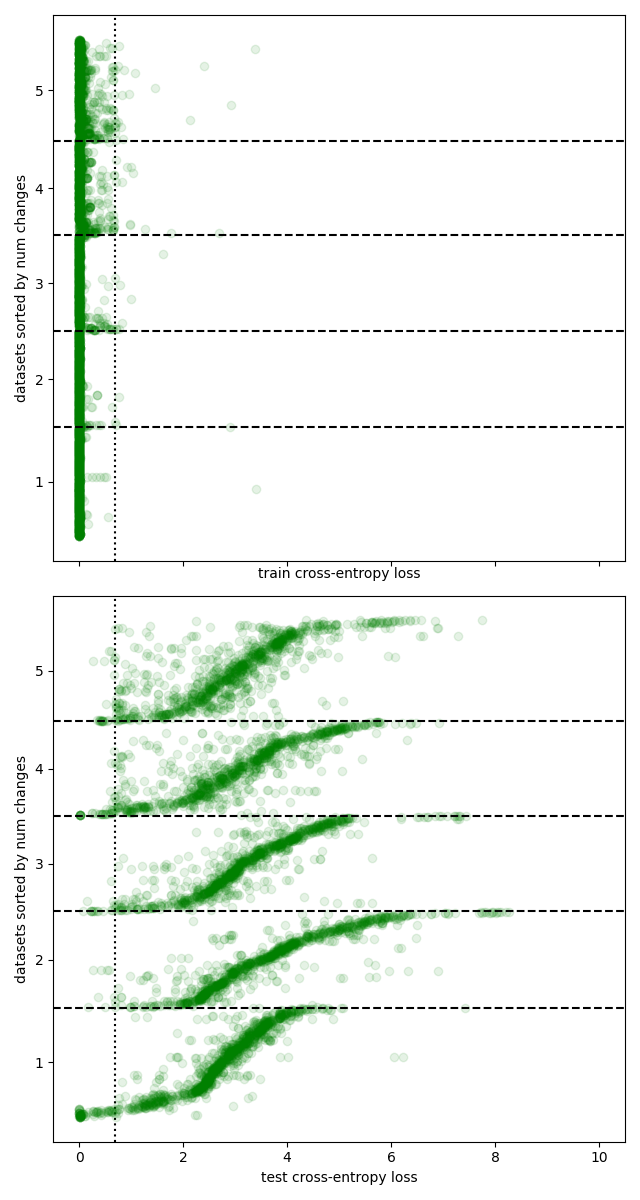}
            \subcaption{GRU}
        \end{minipage} &
        \begin{minipage}[t]{0.3\linewidth}
            \centering
            \includegraphics[width=\linewidth]{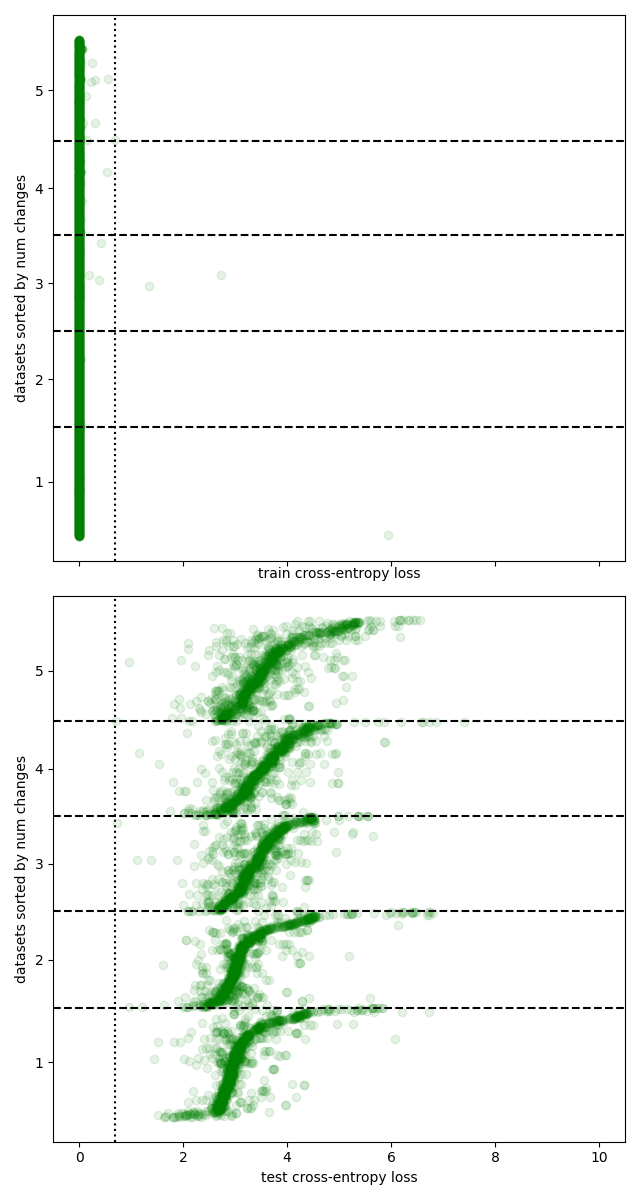}
            \subcaption{Elman RNN}
        \end{minipage}
    \end{tabular}
    \caption{The Scatter plot of the train/test cross-entropies for LSTM, GRU, and Elman RNN with $(\numlayers,\; \hiddensize)\eqspace (1, 200)$. The horizontal dashed line separates the datasets by the number of label changes. Besides, the datasets are also sorted by the median test cross-entropy. The dotted vertical line shows the random baseline loss of $-\ln (\frac{1}{2}) \approx 0.7$.}
    \label{fig:raw_loss_L1}
\end{figure*}

\begin{figure*}[t]
    \begin{tabular}{ccc}
        \begin{minipage}[t]{0.3\linewidth}
            \centering
            \includegraphics[width=\linewidth]{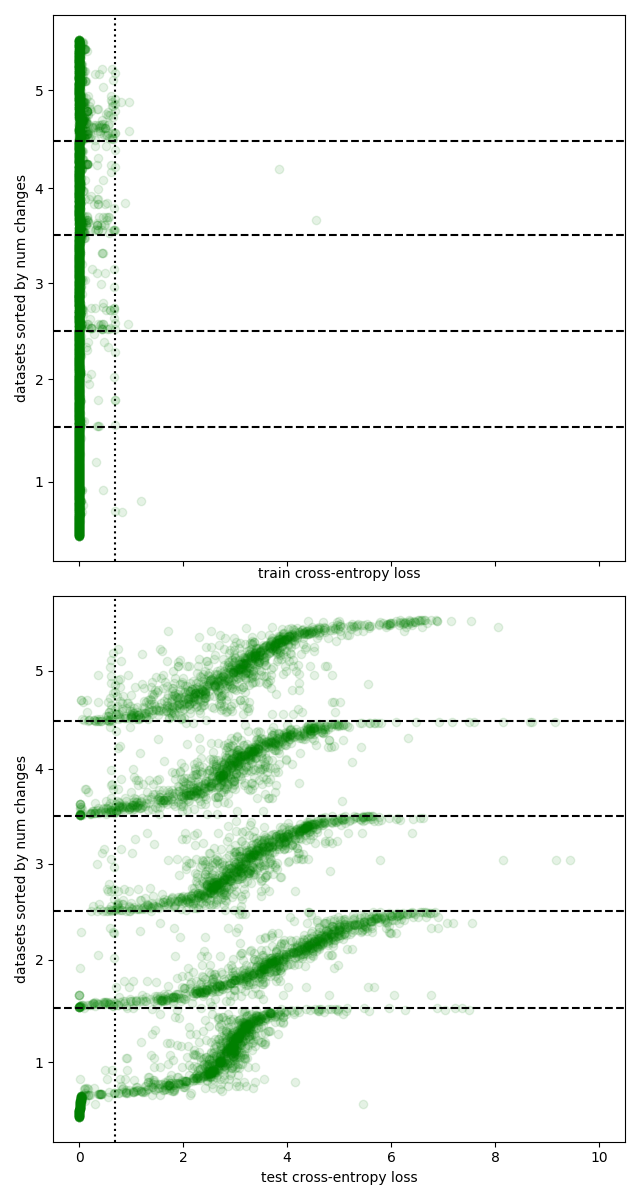}
            \subcaption{LSTM}
        \end{minipage} &
        \begin{minipage}[t]{0.3\linewidth}
            \centering
            \includegraphics[width=\linewidth]{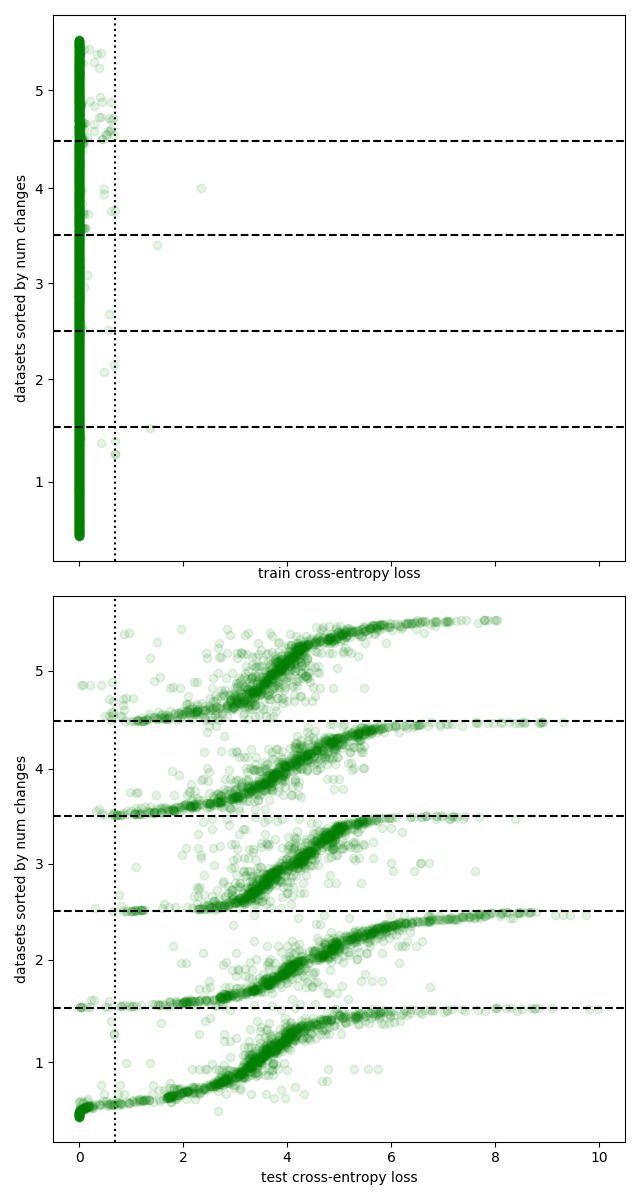}
            \subcaption{GRU}
        \end{minipage} &
        \begin{minipage}[t]{0.3\linewidth}
            \centering
            \includegraphics[width=\linewidth]{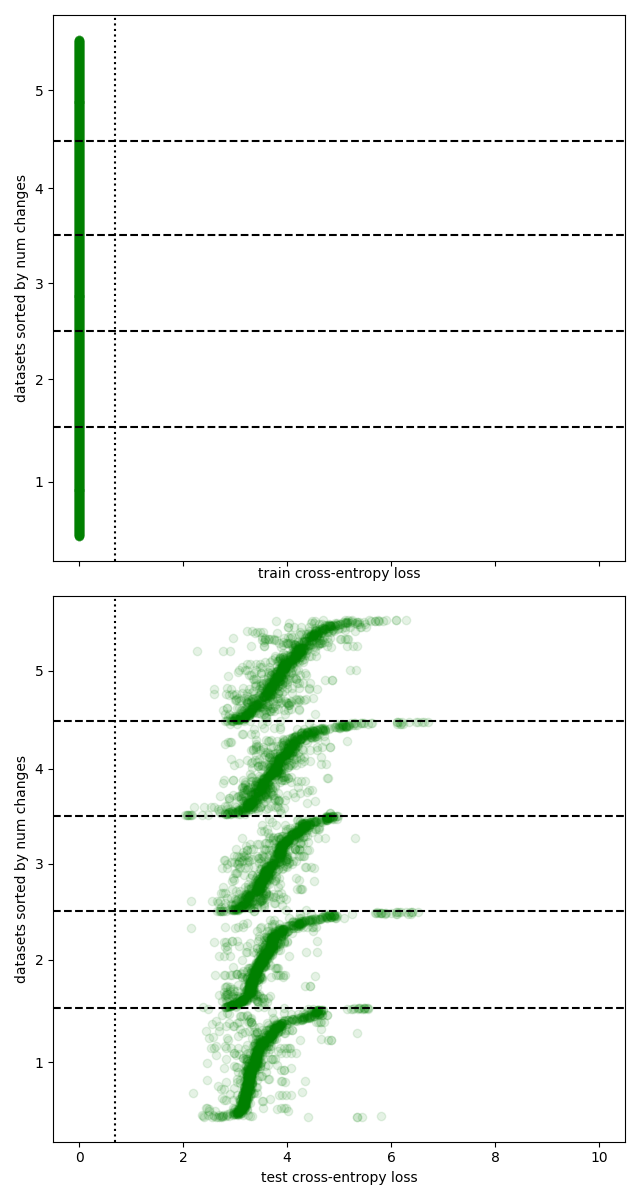}
            \subcaption{Elman RNN}
        \end{minipage}
    \end{tabular}
    \caption{The Scatter plot of the train/test cross-entropies for LSTM, GRU, and Elman RNN with $(\numlayers,\; \hiddensize)\eqspace (2, 200)$. The horizontal dashed line separates the datasets by the number of label changes. Besides, the datasets are also sorted by the median test cross-entropy. The dotted vertical line shows the random baseline loss of $-\ln (\frac{1}{2}) \approx 0.7$.}
    \label{fig:raw_loss_L2}
\end{figure*}

\begin{figure*}[t]
    \begin{tabular}{ccc}
        \begin{minipage}[t]{0.3\linewidth}
            \centering
            \includegraphics[width=\linewidth]{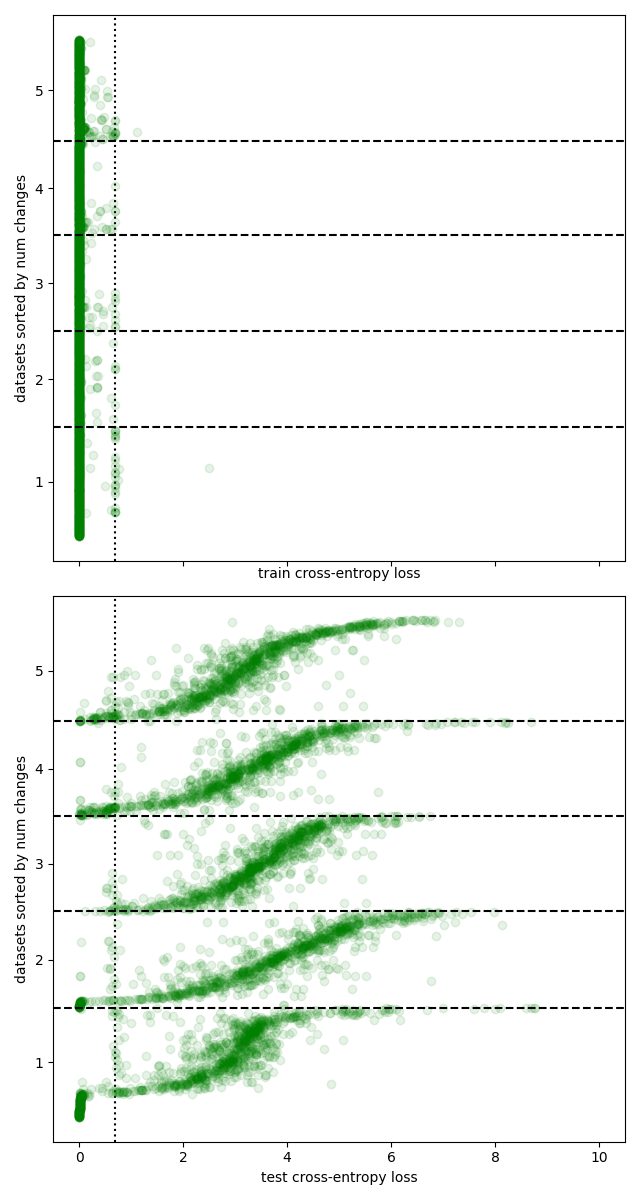}
            \subcaption{LSTM}
        \end{minipage} &
        \begin{minipage}[t]{0.3\linewidth}
            \centering
            \includegraphics[width=\linewidth]{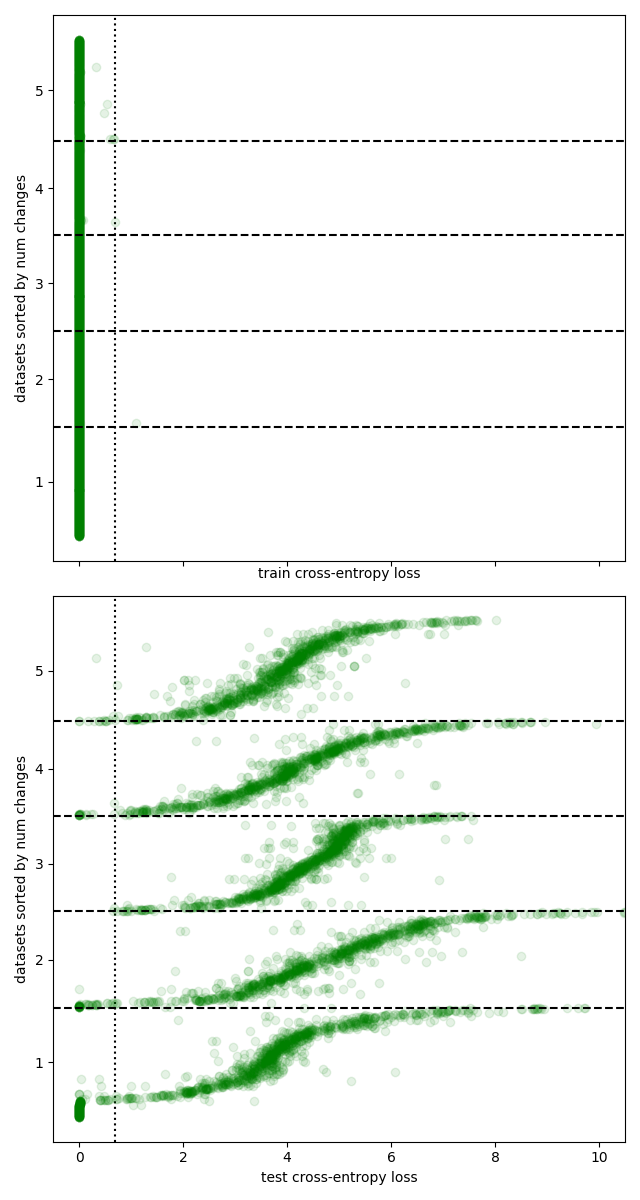}
            \subcaption{GRU}
        \end{minipage} &
        \begin{minipage}[t]{0.3\linewidth}
            \centering
            \includegraphics[width=\linewidth]{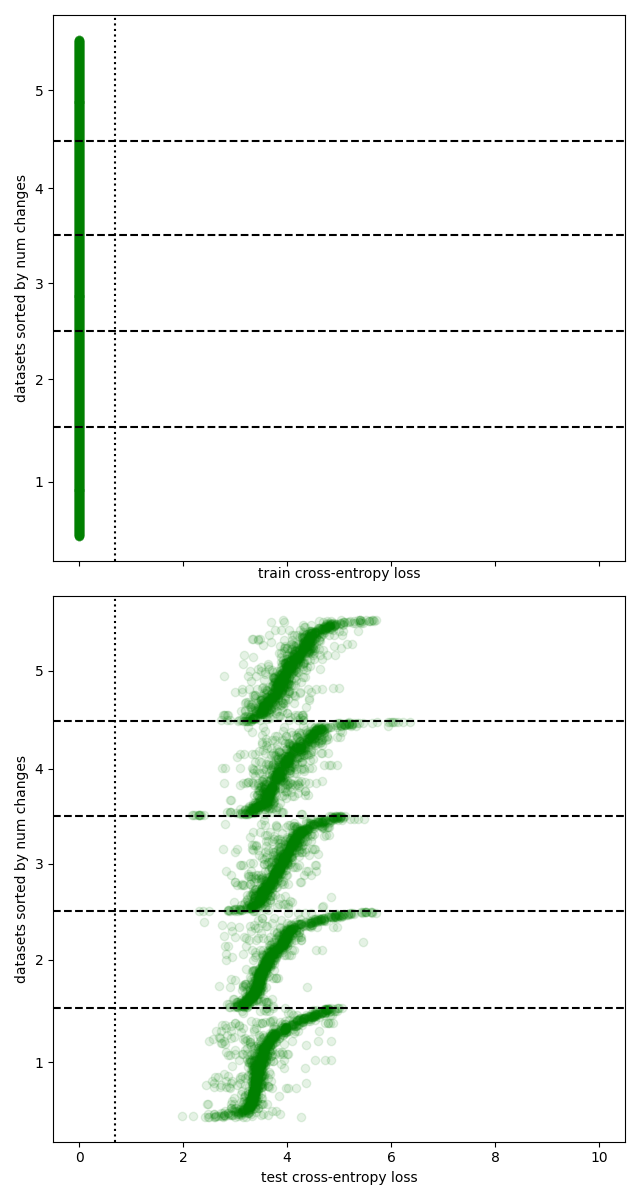}
            \subcaption{Elman RNN}
        \end{minipage}
    \end{tabular}
    \caption{The Scatter plot of the train/test cross-entropies for LSTM, GRU, and Elman RNN with $(\numlayers,\; \hiddensize)\eqspace (3, 200)$. The horizontal dashed line separates the datasets by the number of label changes. Besides, the datasets are also sorted by the median test cross-entropy. The dotted vertical line shows the random baseline loss of $-\ln (\frac{1}{2}) \approx 0.7$.}
    \label{fig:raw_loss_H2000}
\end{figure*}

\begin{figure*}[t]
    \begin{tabular}{ccc}
        \begin{minipage}[t]{0.3\linewidth}
            \centering
            \includegraphics[width=\linewidth]{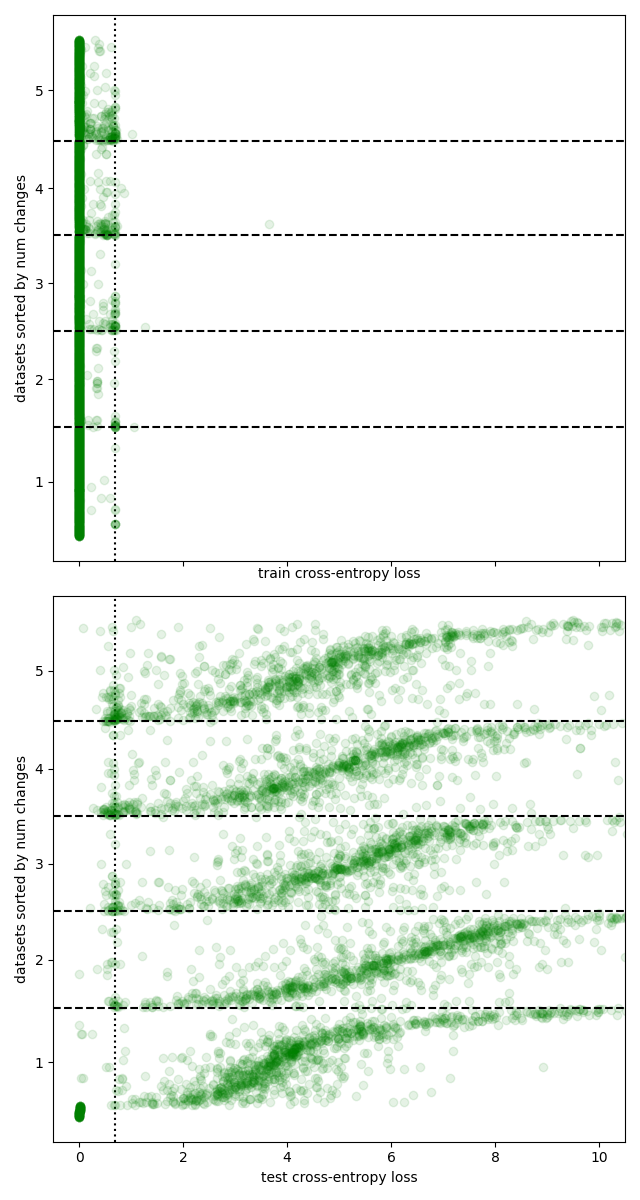}
            \subcaption{LSTM}
        \end{minipage} &
        \begin{minipage}[t]{0.3\linewidth}
            \centering
            \includegraphics[width=\linewidth]{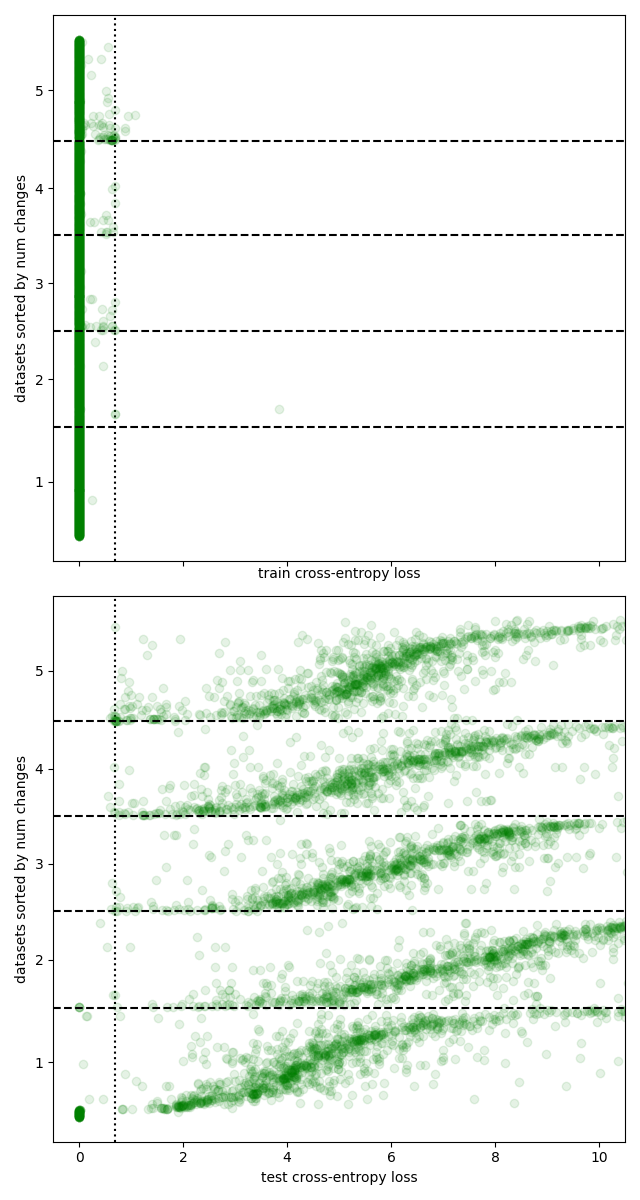}
            \subcaption{GRU}
        \end{minipage} &
        \begin{minipage}[t]{0.3\linewidth}
            \centering
            \includegraphics[width=\linewidth]{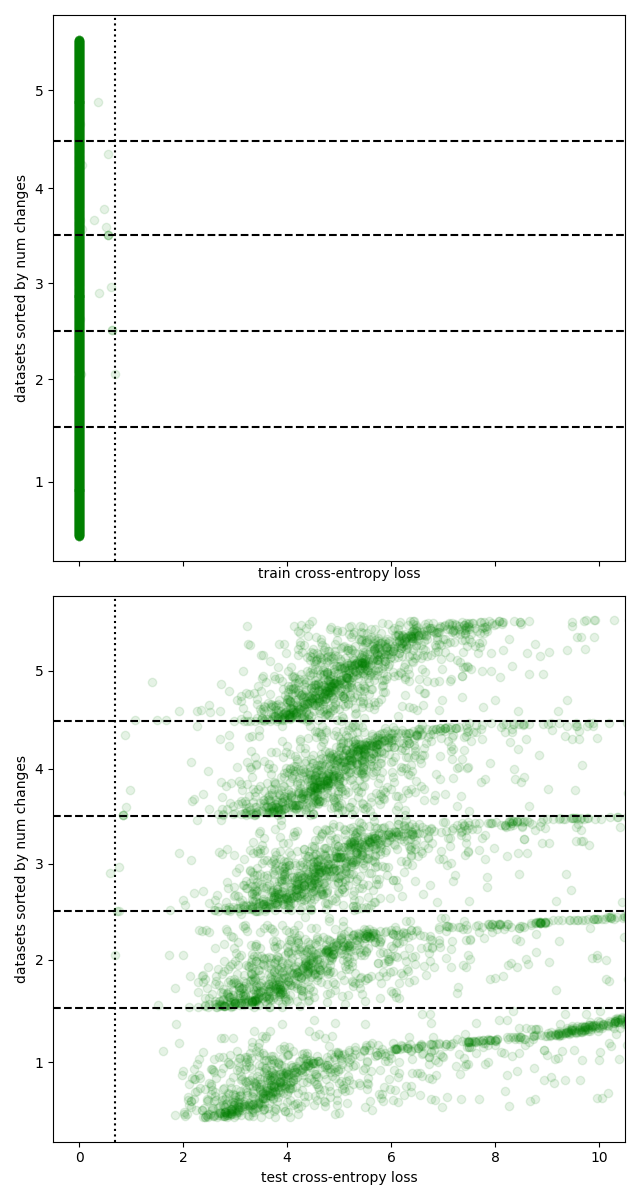}
            \subcaption{Elman RNN}
        \end{minipage}
    \end{tabular}
    \caption{The Scatter plot of the train/test cross-entropies for LSTM, GRU, and Elman RNN with $(\numlayers,\; \hiddensize)\eqspace (2, 2000)$. The horizontal dashed line separates the datasets by the number of label changes. Besides, the datasets are also sorted by the median test cross-entropy. The dotted vertical line shows the random baseline loss of $-\ln (\frac{1}{2}) \approx 0.7$.}
    \label{fig:raw_loss_L3}
\end{figure*}

In \autoref{fig:raw_loss_L1}, \autoref{fig:raw_loss_L2}, \autoref{fig:raw_loss_L3}, and \autoref{fig:raw_loss_H2000}, we show the scatter plot of the train/test cross-entropies for LSTM, GRU, and Ellman RNN for all the settings.
The horizontal dashed line separates the datasets by the number of label changes.
Besides, the datasets are also sorted by the median test cross-entropy.
The dotted vertical line shows the random baseline loss of $-\ln (\frac{1}{2}) \approx 0.7$.

In \autoref{fig:all_loss}, the number of datasets having near-zero test cross-entropy is relatively higher for LSTM and GRU.
For example, from \autoref{fig:raw_loss_L1} (a), \autoref{fig:raw_loss_L2} (a), \autoref{fig:raw_loss_L3} (a), and \autoref{fig:raw_loss_H2000} (a), we can see that the datasets with the near-zero test cross-entropy loss mostly have only 1 label change.
This indicates that LSTM and GRU indeed sometimes learn to naively extend the given labels, but mostly in the extreme case where the datasets have only 1 label change.
However, for Elman RNN, we cannot find such a trend.


\section{Raw Dominant Frequency}\label{appendix:raw_domfreq}
\begin{figure*}[t]
    \begin{tabular}{ccc}
        \begin{minipage}[t]{0.3\linewidth}
            \centering
            \includegraphics[width=\linewidth]{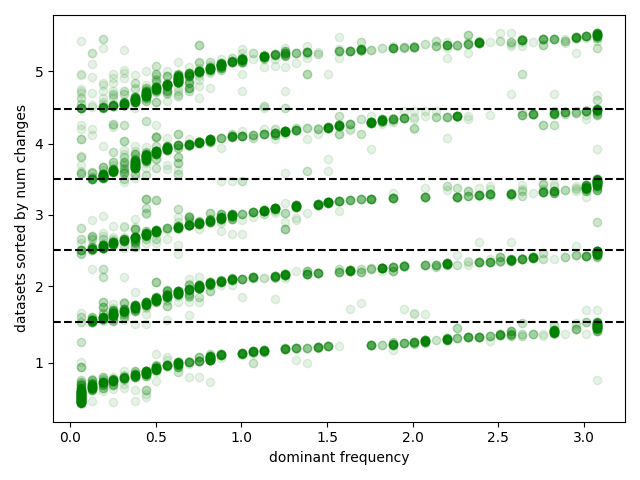}
            \subcaption{LSTM}
        \end{minipage} &
        \begin{minipage}[t]{0.3\linewidth}
            \centering
            \includegraphics[width=\linewidth]{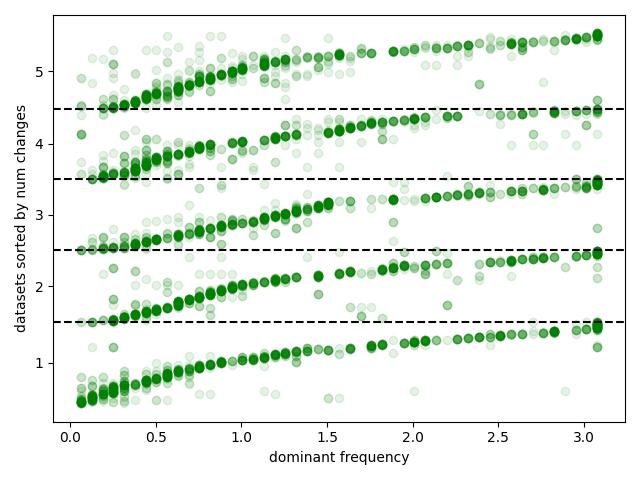}
            \subcaption{GRU}
        \end{minipage} &
        \begin{minipage}[t]{0.3\linewidth}
            \centering
            \includegraphics[width=\linewidth]{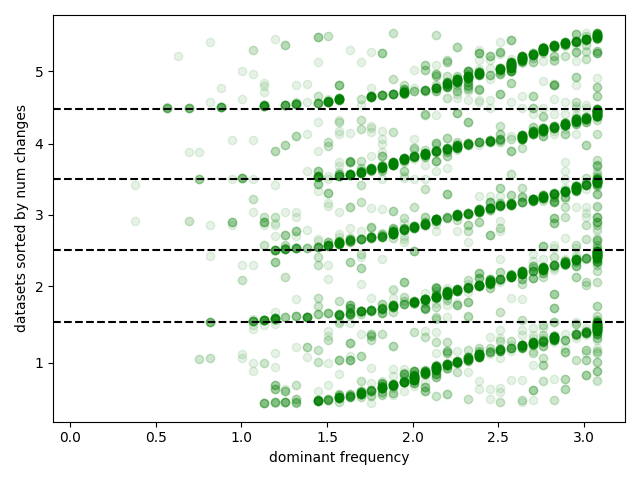}
            \subcaption{Elman RNN}
        \end{minipage}
    \end{tabular}
    \caption{The Scatter plot of the dominant frequencies for LSTM, GRU, and Elman RNN with $(\numlayers,\; \hiddensize)\eqspace (1, 200)$. The horizontal dashed line separates the datasets by the number of label changes. Besides, the datasets are also sorted by the median dominant frequencies.}
    \label{fig:raw_domfreq_L1}
\end{figure*}

\begin{figure*}[t]
    \begin{tabular}{ccc}
        \begin{minipage}[t]{0.3\linewidth}
            \centering
            \includegraphics[width=\linewidth]{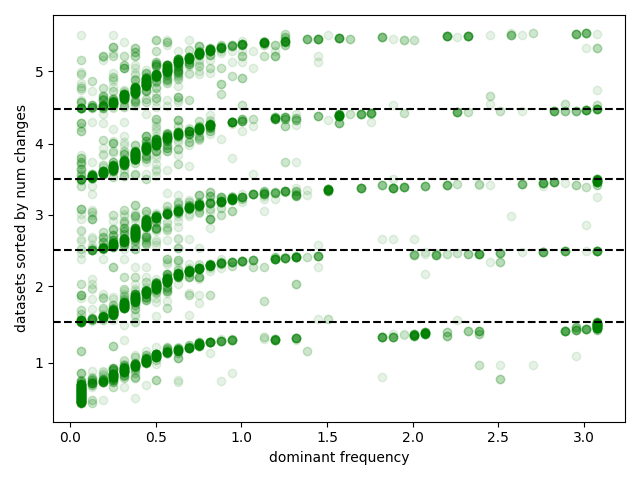}
            \subcaption{LSTM}
        \end{minipage} &
        \begin{minipage}[t]{0.3\linewidth}
            \centering
            \includegraphics[width=\linewidth]{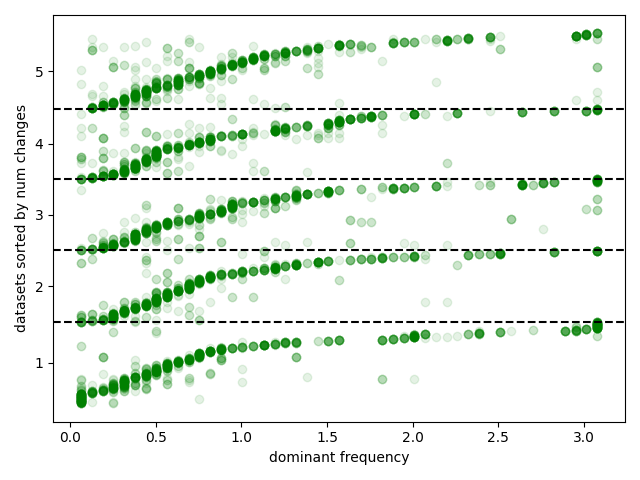}
            \subcaption{GRU}
        \end{minipage} &
        \begin{minipage}[t]{0.3\linewidth}
            \centering
            \includegraphics[width=\linewidth]{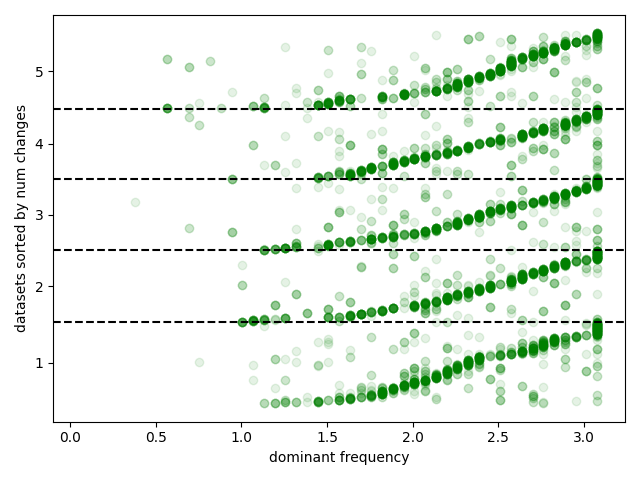}
            \subcaption{Elman RNN}
        \end{minipage}
    \end{tabular}
    \caption{The Scatter plot of the dominant frequencies for LSTM, GRU, and Elman RNN with $(\numlayers,\; \hiddensize)\eqspace (2, 200)$. The horizontal dashed line separates the datasets by the number of label changes. Besides, the datasets are also sorted by the median dominant frequencies.}
    \label{fig:raw_domfreq_L2}
\end{figure*}

\begin{figure*}[t]
    \begin{tabular}{ccc}
        \begin{minipage}[t]{0.3\linewidth}
            \centering
            \includegraphics[width=\linewidth]{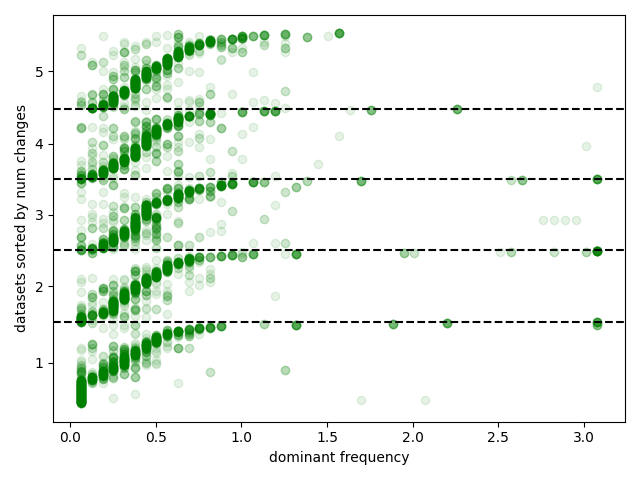}
            \subcaption{LSTM}
        \end{minipage} &
        \begin{minipage}[t]{0.3\linewidth}
            \centering
            \includegraphics[width=\linewidth]{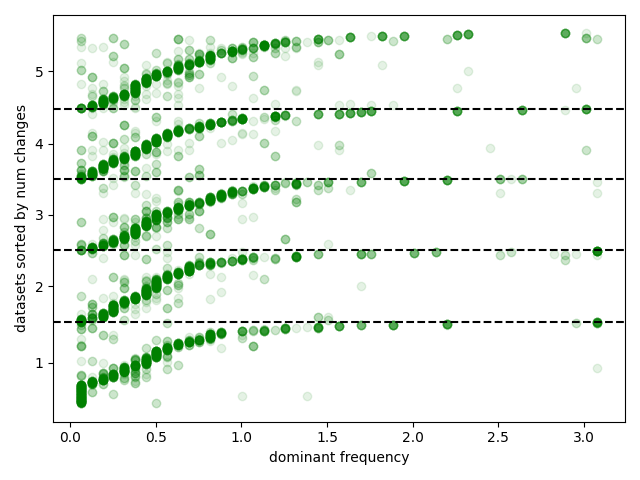}
            \subcaption{GRU}
        \end{minipage} &
        \begin{minipage}[t]{0.3\linewidth}
            \centering
            \includegraphics[width=\linewidth]{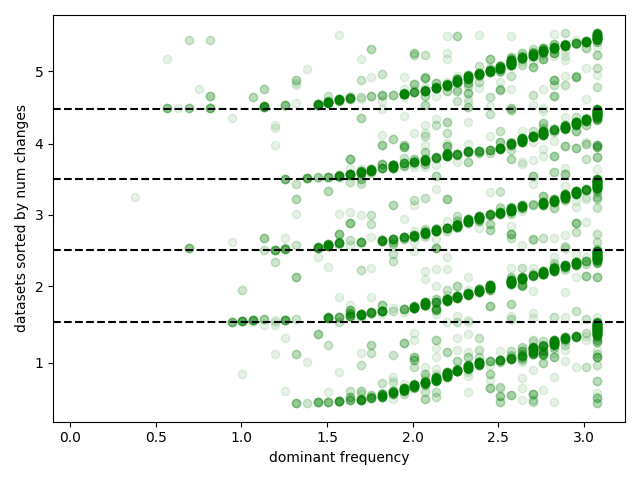}
            \subcaption{Elman RNN}
        \end{minipage}
    \end{tabular}
    \caption{The Scatter plot of the dominant frequencies for LSTM, GRU, and Elman RNN with $(\numlayers,\; \hiddensize)\eqspace (3, 200)$. The horizontal dashed line separates the datasets by the number of label changes. Besides, the datasets are also sorted by the median dominant frequencies.}
    \label{fig:raw_domfreq_L3}
\end{figure*}

\begin{figure*}[t]
    \begin{tabular}{ccc}
        \begin{minipage}[t]{0.3\linewidth}
            \centering
            \includegraphics[width=\linewidth]{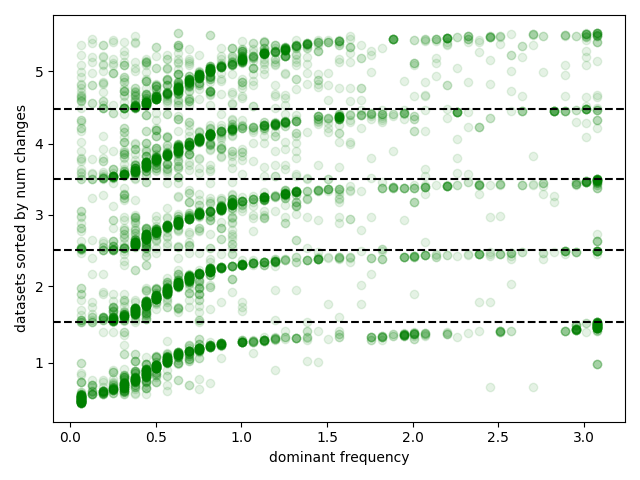}
            \subcaption{LSTM}
        \end{minipage} &
        \begin{minipage}[t]{0.3\linewidth}
            \centering
            \includegraphics[width=\linewidth]{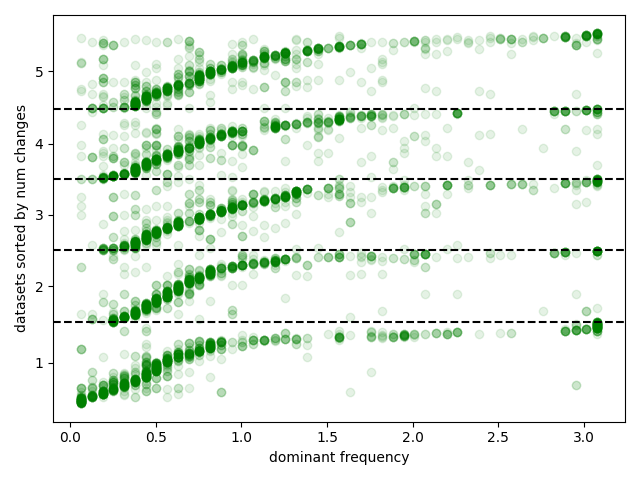}
            \subcaption{GRU}
        \end{minipage} &
        \begin{minipage}[t]{0.3\linewidth}
            \centering
            \includegraphics[width=\linewidth]{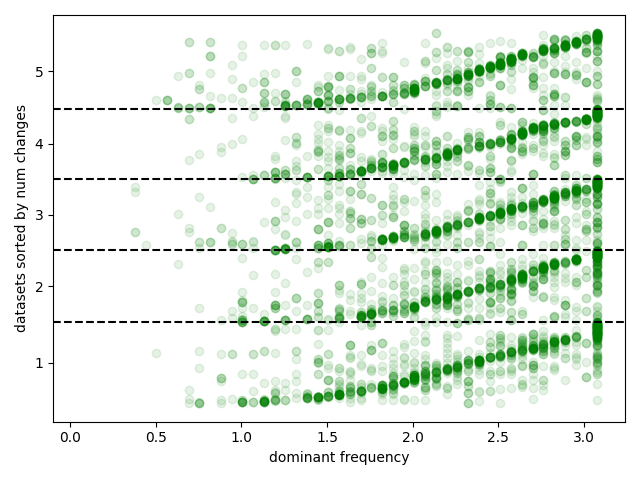}
            \subcaption{Elman RNN}
        \end{minipage}
    \end{tabular}
    \caption{The Scatter plot of the dominant frequencies for LSTM, GRU, and Elman RNN with $(\numlayers,\; \hiddensize)\eqspace (2, 2000)$. The horizontal dashed line separates the datasets by the number of label changes. Besides, the datasets are also sorted by the median dominant frequencies.}
    \label{fig:raw_domfreq_H2000}
\end{figure*}

In \autoref{fig:raw_domfreq_L1}, \autoref{fig:raw_domfreq_L2}, \autoref{fig:raw_domfreq_L3}, and \autoref{fig:raw_domfreq_H2000}, we show the scatter plot of the dominant frequencies for LSTM, GRU, and Ellman RNN for all the settings.
The horizontal dashed line separates the datasets by the number of label changes.
Besides, the datasets are also sorted by the median dominant frequency.

In \autoref{fig:all_loss} (b, c), the number of datasets having the lowest frequency pattern is relatively higher for LSTM and GRU.
We can see that these lowest frequency patterns are mostly restricted to the datasets having only 1 label change (\autoref{fig:raw_domfreq_L1} (a), \autoref{fig:raw_domfreq_L2} (a), \autoref{fig:raw_domfreq_L3} (a), and \autoref{fig:raw_domfreq_H2000} (a)).
This should be consistent with the findings in \autoref{appendix:raw_loss}.
When a model simply learns to extend the labels, its dominant frequency is expected to be near its lowest when there is only one label change in the training dataset since the output sequence contains only one output change in such a case.


\end{document}